\documentclass{article}

    \usepackage[preprint]{neurips_2025}

\usepackage[utf8]{inputenc} 
\usepackage[T1]{fontenc}    
\usepackage[hidelinks,bookmarksopen=true]{hyperref}       
\usepackage{url}            
\usepackage{booktabs}       
\usepackage{amsfonts}       
\usepackage{nicefrac}       
\usepackage{microtype}      
\usepackage{xcolor}         
\PassOptionsToPackage{plain}{natbib}
\usepackage{wrapfig} 
\usepackage{authblk}

\usepackage{amsmath}
\usepackage{amsthm}
\usepackage{tikz}
\usepackage[ruled,vlined,linesnumbered]{algorithm2e}
\usepackage{xcolor}
\usepackage{comment,bm,graphicx,multirow}
\graphicspath{{./fig/}}

\def\vs{\emph{vs.~}}
\def\ie{\emph{i.e.,~}}

\def\etal{{\em et al.}}

\newcommand{\pair}[1]{$[$\tikz[baseline=(base)]{\node[inner sep=0pt](base)at(0,0){};
\begin{scope}[yshift=-0.5ex]
\fill[#1,rounded corners=1pt](0,0) rectangle (0.3,0.3);  
\fill[#1,rounded corners=1pt](0.5,0) rectangle (0.4,0.3); \end{scope}}$]$}
\definecolor{blue_prompt}{HTML}{2E75B6}
\definecolor{black_prompt}{HTML}{767171}

\ifdefined \GramaCheck
  \newcommand{\CheckRmv}[1]{}
  \newcommand{\figref}[1]{Figure 1}
  \newcommand{\tabref}[1]{Table 1}
  \newcommand{\secref}[1]{Section 1}
  \renewcommand{\eqref}[1]{Equation 1}
  \newcommand{\algref}[1]{Algorithm 1}
\else
  \newcommand{\CheckRmv}[1]{#1}
  \newcommand{\figref}[1]{Fig.~\ref{#1}}
  \newcommand{\tabref}[1]{Tab.~\ref{#1}}
  \newcommand{\secref}[1]{Sec.~\ref{#1}}
  \renewcommand{\eqref}[1]{Eq.~(\ref{#1})}
  \newcommand{\algref}[1]{Alg.~\ref{#1}}
\fi

\newcommand{\tabSpace}{\vspace{6pt}}  
\newcommand{\tabFormat}{\centering \renewcommand{\arraystretch}{1.05}}  

\title{PrePrompt: Predictive prompting for class incremental learning}

\author[1]{Libo~Huang}
\author[1]{Zhulin~An}
\author[1]{Chuanguang~Yang}
\author[1]{Boyu~Diao}
\author[1]{Fei~Wang}
\author[2]{Yan~Zeng}
\author[3]{Zhifeng~Hao}
\author[1]{Yongjun~Xu}

\affil[1]{Institute of Computing Technology, Chinese Academy of Sciences}
\affil[2]{Department of Mathematics and Statistics, Beijing Technology and Business University}
\affil[3]{College of Science, Shantou University}

\begin{document}

\maketitle

\begin{abstract}
Class Incremental Learning (CIL) based on pre-trained models offers a promising direction for open-world continual learning. Existing methods typically rely on correlation-based strategies, where an image's classification feature is used as a query to retrieve the most related key prompts and select the corresponding value prompts for training. However, these approaches face an inherent limitation: fitting the entire feature space of all tasks with only a few trainable prompts is fundamentally challenging. We propose Predictive Prompting (PrePrompt), a novel CIL framework that circumvents correlation-based limitations by leveraging pre-trained models' natural classification ability to predict task-specific prompts. Specifically, PrePrompt decomposes CIL into a two-stage prediction framework: task-specific prompt prediction followed by label prediction. While theoretically appealing, this framework risks bias toward recent classes due to missing historical data for older classifier calibration. PrePrompt then mitigates this by incorporating feature translation, dynamically balancing stability and plasticity. Experiments across multiple benchmarks demonstrate PrePrompt's superiority over state-of-the-art prompt-based CIL methods.  Code available at \href{github.com/libo-huang/preprompt}{github.com/libo-huang/preprompt}.
\end{abstract}


\section{Introduction} \label{sec:intro}
\begin{wrapfigure}{r}{0.55\textwidth}
    \centering
    \vspace{-44px}
    \includegraphics[width=\linewidth]{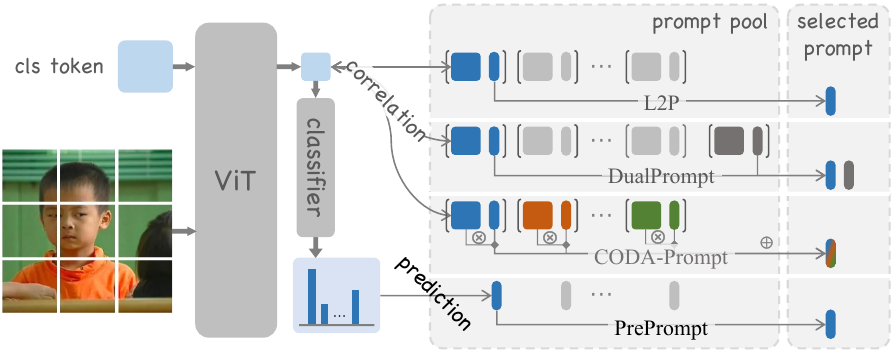}
    \vspace{-15px}
    \caption{Comparison of existing prompt-based CIL methods and our PrePrompt approach. 
    Existing prompt-based CIL methods typically utilize prompts as key-value pairs (\ie \protect\pair{blue_prompt}) and select prompts based on feature-key correlation. Specifically, L2P employs k-nearest neighbors for selection, DualPrompt introduces a set of general-purpose prompts (\ie \protect\pair{black_prompt}), and CODA-Prompt combines all prompts with weights determined by their correlation. 
    In contrast, the proposed PrePrompt simplifies this structure by using a single value prompt and selects the final prompt based on the initial prediction.}
    \label{fig:prompts}
    \vspace{-20px}
\end{wrapfigure}
Deep learning models have demonstrated remarkable success in various applications through conventional training paradigms that utilize pre-collected datasets~\cite{lecun2015deep,xu2021artificial}. 
However, real-world deployment scenarios introduce a critical challenge: constrained by privacy preservation requirements and finite storage capacity, artificial intelligence systems must progressively acquire knowledge from sequentially arriving tasks~\cite{wang2024comprehensive}.
In response to these constraints, Class-Incremental Learning (CIL) has emerged as an essential machine learning paradigm, enabling models to continuously expand their classification capabilities while maintaining previously acquired knowledge~\cite{masana2022class,zhou2024class}. 
The primary obstacle in CIL frameworks is catastrophic forgetting, a phenomenon where direct optimization on new task data induces exponential performance degradation on previously learned classes~\cite{mccloskey1989catastrophic}.

\newcommand{\addFig}[1]{{\includegraphics[width=.245\textwidth]{#1}}}
\begin{figure*}[t]
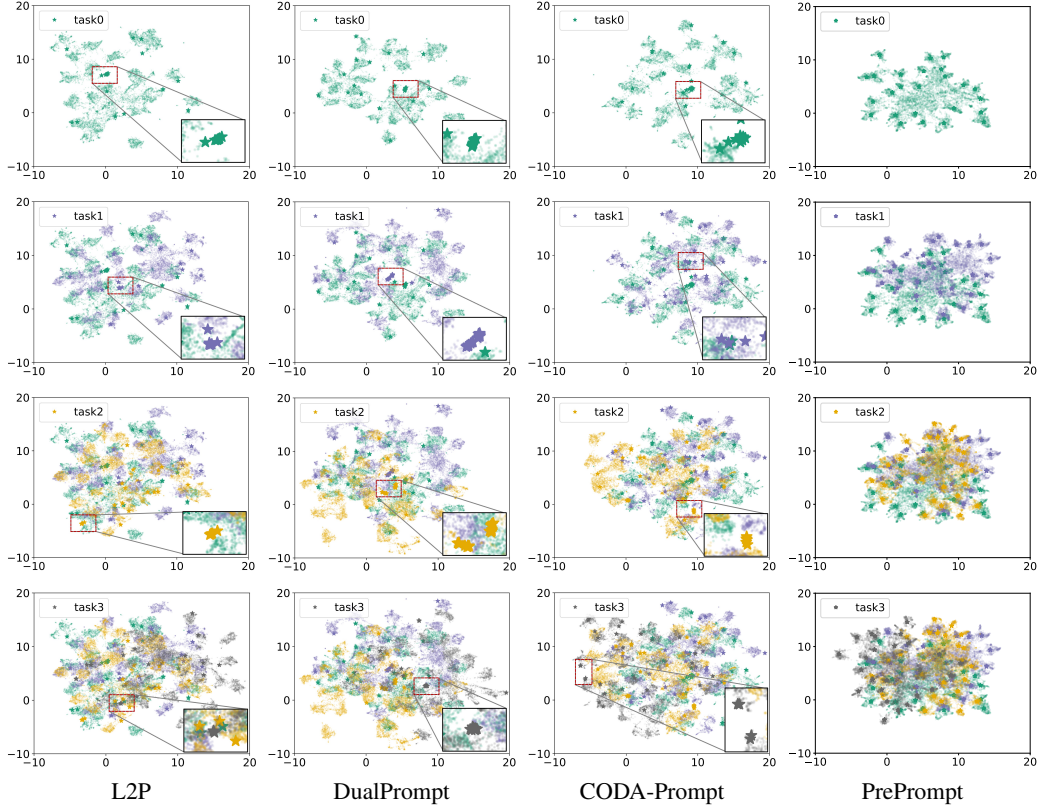

    \centering
    \small
    \setlength{\tabcolsep}{0.2mm}
    \begin{tabular}{cccc}
    \addFig{tsne/l2p_tsne_0.jpg} &
    \addFig{tsne/dualprompt_tsne_0.jpg}&
    \addFig{tsne/codaprompt_tsne_0.jpg}&
    \addFig{tsne/preprompt_tsne_0.pdf}
    \\
    \addFig{tsne/l2p_tsne_1.jpg} &
    \addFig{tsne/dualprompt_tsne_1.jpg}&
    \addFig{tsne/codaprompt_tsne_1.jpg}&
    \addFig{tsne/preprompt_tsne_1.pdf}
    \\    
    \addFig{tsne/l2p_tsne_2.jpg} &
    \addFig{tsne/dualprompt_tsne_2.jpg}&
    \addFig{tsne/codaprompt_tsne_2.jpg}&
    \addFig{tsne/preprompt_tsne_2.pdf}
    \\
    \addFig{tsne/l2p_tsne_3.jpg} &
    \addFig{tsne/dualprompt_tsne_3.jpg}&
    \addFig{tsne/codaprompt_tsne_3.jpg}&
    \addFig{tsne/preprompt_tsne_3.pdf}
    \\
    L2P & DualPrompt & CODA-Prompt & PrePrompt \\
    \end{tabular}
    \caption{
    T-SNE visualization of feature and prompt distributions on CIFAR-100. The plot compares key-prompt embeddings (large stars, color-coded by task) and feature embeddings (small dots) across four sequential tasks, each containing 25 classes. Since PrePrompt does not use key-prompts, we visualize the class feature means instead, where the class IDs are inferred by the prompt classifier. L2P, DualPrompt, and CODA-Prompt exhibit spatial overlap and clustering in their key-prompt embeddings (with some regions zoomed out), indicating inter-task interference. PrePrompt’s embeddings remain well-separated and strategically distributed in distinct regions. This highlights PrePrompt’s superior task-specific decoupling and robustness against catastrophic forgetting in prompt-based class incremental learning.
    }
    \vspace{-15px}
    \label{fig:tsne}
\end{figure*}

While conventional CIL methodologies often require training models from scratch for each new task, recent advancements have shifted toward leveraging pre-trained models (PTMs) as foundational knowledge repositories \cite{wang2022learning, jia2022visual}. 
These PTM-based frameworks exploit the generalizable knowledge encoded in pre-trained transformers, significantly enhancing learning efficiency in CIL scenarios. 
Among these advancements, prompt-based CIL methods have emerged as a promising direction due to their parameter efficiency and adaptability \cite{lester2021power}.
By freezing the transformer backbone and appending lightweight prompt modules, these methods guide feature adaptation while requiring less than $1\%$ of the total model parameters to be updated \cite{zhou2024class, ijcai2024p924}.
As illustrated in \figref{fig:prompts}, state-of-the-art methods such as Learning to Prompt (L2P)~\citep{wang2022learning}, DualPrompt~\cite{wang2022dualprompt}, and CODA-Prompt~\citep{smith2023coda} organize prompts as key-value pairs, where image classification features serve as queries to select the most relevant key prompts, which in turn determine the corresponding value prompts for training. 
L2P selects prompts using a k-nearest neighbors mechanism but often fails to achieve sufficient task-specific feature separation~\citep{wang2022learning}. 
DualPrompt introduces general-purpose prompts to mitigate this issue but achieves only partial separation~\cite{wang2022dualprompt}.
Similarly, CODA-Prompt attempts to combine all prompts with weighted correlations using an attention mechanism~\citep{smith2023coda}.
Although these methods have pushed performance levels close to the theoretical upper bounds of conventional CIL, they rely heavily on predefined correlations between classification features (queries) and key prompts.
This reliance limits their ability to represent the complex and diverse feature spaces of multiple tasks. 
As tasks increase in number or diversity, overlaps between prompts associated with different tasks become inevitable, reducing task separability and diminishing performance gains.
These shortcomings are evident in \figref{fig:tsne}, where the prompts from L2P, DualPrompt, and CODA-Prompt exhibit varying degrees of spatial overlap, with their embeddings partially clustered together, indicating potential interference between tasks due to insufficient prompt separation.

In this paper, we propose PrePrompt, a novel prediction-based prompt selection strategy that simplifies the key-value pair framework used in prior methods. 
As shown in \figref{fig:prompts}, PrePrompt eliminates the need for a separate key-value structure and reduces reliance on correlation-based queries.
Instead of relying on predefined correlations, PrePrompt leverages the model's initial prediction capability to identify the most appropriate prompt for each task and subsequently predicts the final label using the selected prompt.
To achieve an optimal balance between stability and plasticity, PrePrompt incorporates a feature translation strategy across these two prediction stages.
As demonstrated in \figref{fig:tsne}, PrePrompt achieves two significant improvements over existing methods.
First, it ensures that prompts are distributed across distinct feature subspaces, effectively reducing task overlap and mitigating interference between tasks. 
Second, it avoids the partial clustering observed in previous methods, thereby enhancing task-specific adaptability and improving the model's ability to handle diverse and complex feature spaces.
These advancements collectively address the limitations of existing prompt-based CIL methods, offering a more robust and scalable solution for continual learning scenarios.
In summary, our contributions are threefold: (1) We identify and analyze the limitations of existing prompt-based CIL methods, particularly their reliance on predefined correlations and insufficient task separability; (2) We propose PrePrompt, a prediction-based prompt selection strategy that eliminates the need for key-value pairs and leverages feature translation to improve task-specific adaptability; and (3) We demonstrate through extensive experiments that PrePrompt achieves superior performance in reducing task overlap and enhancing task separability, outperforming state-of-the-art methods on benchmark datasets.

\vspace{0px}
\section{Related work} \label{sec:related}
\subsection{Incremental learning}
\vspace{-5px}
Continual learning enables neural networks to acquire new knowledge while preserving previously learned capabilities, a critical requirement for real-world applications encountering dynamic data streams. 
Early approaches addressed catastrophic forgetting through regularization-based methods that selectively stabilize important parameters \cite{kirkpatrick2017overcoming,zenke2017continual}, exemplified by synaptic intelligence \cite{zenke2017continual} and elastic weight consolidation \cite{kirkpatrick2017overcoming}.
Rehearsal-based strategies later emerged as dominant solutions, maintaining performance through episodic memory buffers \cite{rebuffi2017icarl} or generative replay \cite{shin2017continual}, though these incur storage overhead and privacy risks.
Architectural innovations further advanced the field, introducing task-specific parameters \cite{serra2018overcoming} or dynamically expanding networks \cite{rusu2016progressive} to isolate conflicting knowledge.
Among various incremental learning scenarios, class-incremental learning (CIL) presents unique challenges by requiring unified prediction across all encountered classes without task identity clues \cite{chaudhry2018efficient}.
Recent CIL methods emphasize balanced feature representations through contrastive learning \cite{cha2021co2l} or prototype alignment \cite{liu2020generative}.
While these architectural and algorithmic innovations have advanced incremental learning, they typically require extensive parameter updates or explicit memory storage. 
This motivates the recent exploration of parameter-efficient prompt-tuning techniques that leverage frozen pre-trained models as knowledge anchors, effectively bridging the gap between incremental adaptability and computational efficiency.

\vspace{-5px}
\subsection{Prompting for incremental learning}
\vspace{-5px}
Building on parameter-efficient tuning paradigms from NLP, prompt-based CIL methods have emerged as a lightweight alternative to traditional incremental learning approaches.
By prepending learnable prompt parameters to frozen pre-trained vision transformers (ViTs), methods like L2P~\cite{wang2022learning} and DualPrompt~\cite{wang2022dualprompt} pioneer prompt pools and task-specific retrieval mechanisms, preserving pre-trained knowledge while adapting to new tasks.
Building on this, CODA-Prompt \cite{smith2023coda} introduces attention-based prompt composition, dynamically fusing prompts through learned similarity weights.
The paradigm further evolves with generative variants like DAP \cite{jung2023generating}, which synthesizes instance-specific prompts via hypernetworks, reducing reliance on fixed pool sizes.\footnote{It is important to note that DAP leaks task identity information during testing by exploiting \textit{non-i.i.d.} test batches, where task IDs are inferred through majority voting. 
This approach violates the fundamental \textit{i.i.d.} evaluation assumption and leads to an unfair comparison, as evidenced by its performance collapsing under shuffled samples or a batch size of 1. Similar findings are independently corroborated by \citep{zhou2024class}.}
Cross-modal extensions such as AttriCLIP \cite{wang2023attriclip} integrate text-guided prompts with vision-language models. 

However, these approaches face a fundamental limitation: fitting the entire feature space of all tasks with a limited set of trainable key prompts is inherently challenging. 
Empirical studies~\cite{zhou2024revisiting} show diminishing returns as task sequences grow, with fixed-size prompt pools failing to mitigate interference between old and new concepts. Unlike existing methods that rely on predefined feature-key prompt correlations to predict the value prompts, our work bypasses this challenge by leveraging the model's inherent classification ability for prompt prediction.

\section{Method}
\subsection{Problem definition}
For continual learning of sequentially arrived $T$ tasks,
each task $i \in \{1,\cdots,T\}$ is represented by the dataset $\mathcal{D}_{i}=\{\mathcal{X}_{i}, \mathcal{Y}_i\}$.
Here, $\mathcal{X}_{i}=\bigcup_{j}\mathcal{X}_{i,j}$ and $\mathcal{Y}_i=\{\bigcup_{j}\mathcal{Y}_{i,j}\}$ are the sample and label sets of the task $i$, respectively; $j \in \left\{1,\cdots,\left|\mathcal{Y}_i\right|\right\}$ indicates the $j$-th class in task $i$; and $|\cdot|$ denotes the cardinal of a set.
Class incremental learning (CIL) assumes the tasks are disjoint, \ie $\mathcal{Y}_i \cap \mathcal{Y}_{i'} =\emptyset$, $\forall i \neq i'$.
Mathematically, consider a sequence of tasks, $\mathcal{D} = \left\{\mathcal{D}_1, \dots, \mathcal{D}_{T}\right\}$, and a pre-trained model parameterized by $\theta$.
After incrementally training the model up to task $t$, the goal of CIL is as follows:
Given any test sample from the learned tasks, $\bm{x} \in \bigcup_{i,j}\mathcal{X}_{i,j}$, the model $\theta$ should predict its label $y \in \{\bigcup_{i,j}\mathcal{Y}_{i,j}\}$ with $P_{\theta}(y\mid\bm{x})$, where $i\in\{1,\cdots,T\}$ and $j \in \{1,\cdots,\left|\mathcal{Y}_i\right|\}$.

\subsection{Predictive prompting (PrePrompt)}

\subsubsection{CIL decomposition with prompt} \label{subsub:cil}
We first decompose the CIL problem into a two-stage prediction framework as,\footnote{For easy presentation, we take the label prediction and prompt prediction related parameters all in the $\theta$.}
\begin{align}\label{eq:p_yx}
    P_{\theta}\left(y\mid\bm{x}\right)=
    P_\theta\left(y \mid \bm{p},\bm{x} \right)
    P_\theta\left(\bm{p} \mid \bm{x} \right),
\end{align}
where the predicted prompt, $\bm{p}$, is expected equal to the $i$-th task-specific prompts set in the prompt pool $\mathcal{P}$, \ie $\bm{p}=\mathcal{P}_{i}\subset\mathcal{P}$. 
This separate prediction is self-consistent; a detailed proof is given below.
\begin{proof}
According to the Bayes-theorem~\citep{koller2009probabilistic}, we have,
\begin{align} \label{eq:p_ypx}
    P_\theta\left(y \mid \bm{p}, \bm{x} \right)=
    \frac{P_{\theta }(\bm{p} \mid y ,\bm{x}) P_{\theta }(y \mid \bm{x})}
    {P_{\theta }(\bm{p} \mid \bm{x})}.
\end{align}
Note that, given the label, $y$, the prediction of $\bm{p}$ becomes deterministic, \ie $P_{\theta }(\bm{p} \mid y ,\bm{x})=1$. Substituting this into \eqref{eq:p_ypx}, we obtain,
\begin{align}\nonumber
    P_\theta\left(y \mid \bm{p}, \bm{x} \right)=
    \frac{P_{\theta }(y \mid \bm{x})}
    {P_{\theta }(\bm{p} \mid \bm{x})},
\end{align}
which is same as \eqref{eq:p_yx}.
\end{proof}
We propose a novel prompt-based CIL method, Predictive Prompting (PrePrompt), to perform the above two-stage prediction, \ie prompt and label predictions.

\subsubsection{PrePrompt framework} \label{subsubsec:preprompt}
\begin{figure*}[!t]
  \centering
  \small
  \setlength{\tabcolsep}{15pt}
  \begin{tabular}{cc}
    \includegraphics[width=0.48\textwidth]{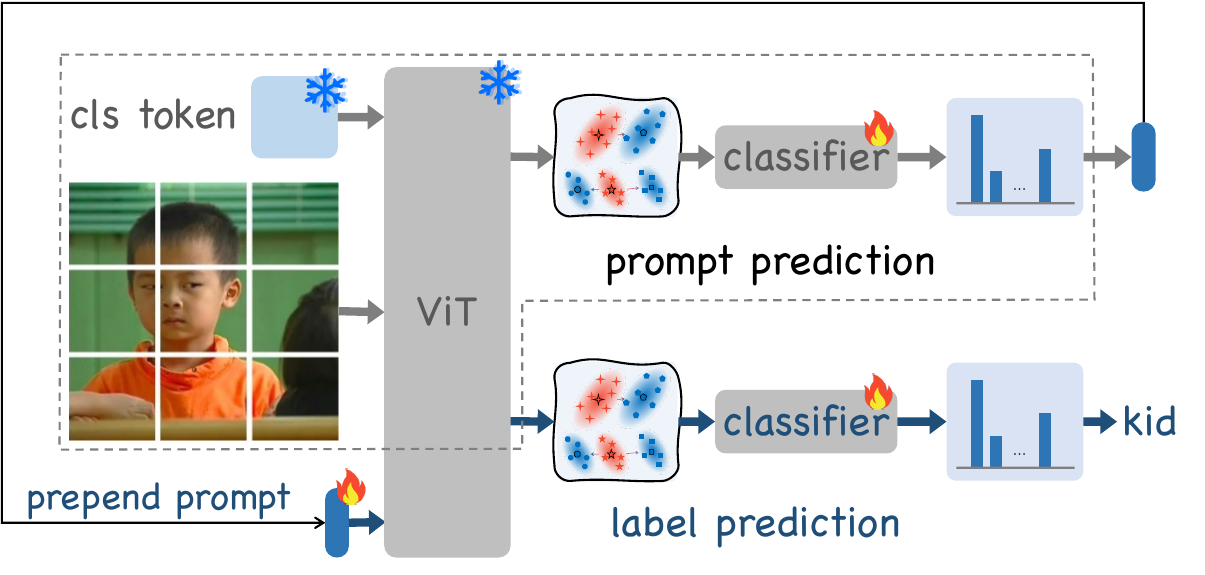} &
    \includegraphics[width=0.40\textwidth]{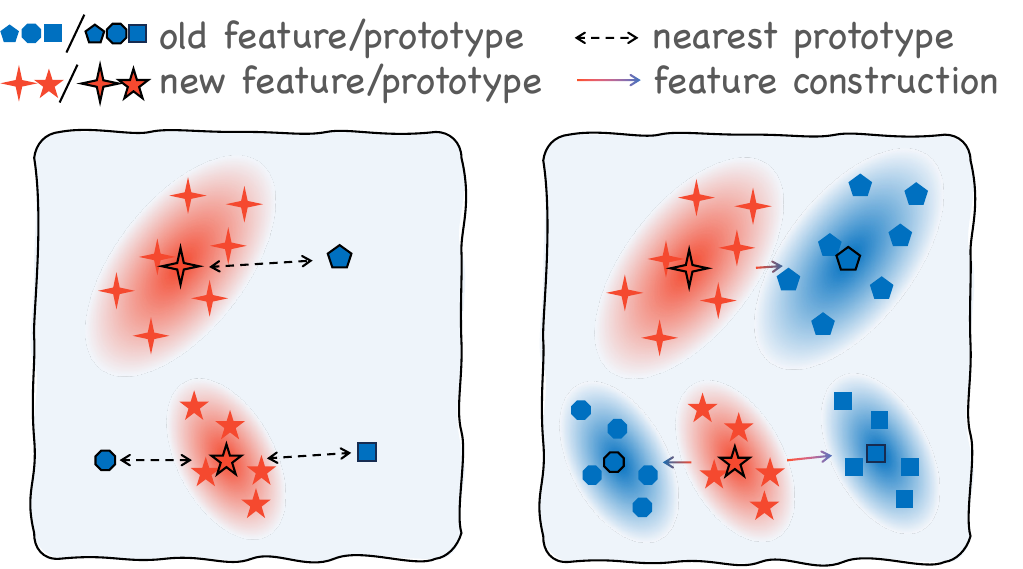} 
    \\
   (a) Framework of PrePrompt & (b) Feature translation \\
  \end{tabular}
  \caption{(a) The PrePrompt framework operates through two sequential prediction stages: prompt prediction and label prediction. During prompt prediction, a pre-trained ViT is utilized to train a prompt classifier, which then predicts task-specific prompts for training a new label classifier to perform final image classification. Both stages leverage (b) feature translation, where the nearest new prototype is identified for each retained old prototype from prior tasks. The old features are constructed by transferring features from their nearest new prototype, followed by alignment to the original old prototype.}
  \label{fig:method}
\end{figure*}

As illustrated in \figref{fig:method}(a), the Preprompt consists of two prediction stages: prompt prediction and label prediction.
In each stage, the vision transformer (ViT) transforms the original image, $\bm{x}\in\mathbb{R}^{H\times W\times C}$, into a feature representation, $\bm{f}\in\mathbb{R}^{D}$. 

The ViT first reshapes $\bm{x}$ into a sequence of $HW/P^2$ flatted patches, $\bm{x_s}=\left[\bm{x}_{s_1},\cdots,\bm{x}_{s_{HW/P^2}}\right]$, where $\bm{x}_{s_i}\in\mathbb{R}^{P\times P\times C}$.
These patches are then mapped from $P\times P\times C$ to $D$ dimensions with a pre-trained embedding projection.
Here, $C$ represents the number of channels in the image, $D$ is the embedding dimension, and $(H, W)$ and $(P, P)$ denote the resolutions of the image and the patches, respectively~\citep{vaswani2017attention,dosovitskiy2020image}.
The pre-trained class token, $\bm{cls}$, along with the position embedding, $\theta_{posE}$, transforms the patches into a representation, $\bm{h}'\in\mathbb{R}^{(HW/P^2+1)\times D}$, as follows:
\begin{align}
    \bm{h}' & = Con(\bm{cls}, \theta_{ProjE}(\bm{x_s}))+\theta_{posE}. \notag 
\end{align}
The ViT comprises several stacked multi-head self-attention (MSA) blocks. 
Each MSA bloack transforms $\bm{h}'$ into a representation, $\bm{h}\in\mathbb{R}^{(HW/P^2+1)\times D}$.
This process begins by linearly projecting $\bm{h}'$ into query ($\bm{h}_Q$), key ($\bm{h}_K$), and value ($\bm{h}_V$) representations. 
These representations are then divided into $m$ equal-length segments, and each head's representation, $\bm{h}''_i$ for  $i=1,...,m$, is computed through its corresponding attention operation:
\begin{align}
    \bm{h}''_i & = \mathrm{softmax}\left(\frac{\bm{h}_{Q_i}\theta_{Q_i}\cdot (\bm{h}_{K_i}\theta_{K_i})^T}{\sqrt{D/m}}\right)\cdot \bm{h}_{V_i}\theta_{V_i}. \notag 
\end{align}
By concatenating all $m$ $\bm{h}''_i$ representations and applying the multi-head projection, $\theta_O$, the final MSA representation, $\bm{h}$, is obtained:
\begin{align} \label{eq:h}
    \bm{h}  & = \mathrm{MSA}(\bm{h}_Q, \bm{h}_K, \bm{h}_V) = Con(\bm{h}''_1,...,\bm{h}''_m)\cdot\theta_O,
\end{align}
where all variables are summarized in~\tabref{tab:var_def} in Appendix~\ref{app:var}.

\noindent
\textbf{Prompt prediction stage.} During this stage, we utilize the inherent classification capabilities of pre-trained models to predict task-specific prompts, $\bm{p}=\mathcal{P}_{i}$.
\textit{In the training phase}, the prompt classifier is trained on the extracted image representation, $\bm{f}=\theta_{FeaE}(\bm{x})$, which corresponds to $\bm{cls}$ after all MSA blocks~\citep{dosovitskiy2020image}, by minimizing the cross-entropy loss,
\begin{align} \label{eq:clap}
\min_{\theta_{ClaP}} \left\{ 
    \mathcal{L}_p(\bm{x}, y)\equiv-\sum_{j \in \mathcal{Y}_i} \delta(j, y) \log\left(\theta_{ClaP}\left(\theta_{FeaE}(\bm{x})\right)_j\right) 
\right\},
\end{align}
where $\delta(j, y)$ is the Kronecker delta function, indicating whether the indices $j$ and $y$ are equal, returning $1$ if they are and $0$ otherwise.
Note that the only trainable parameters are located in the prompt classifier.
\textit{In the inference phase}, we can predict the task-specific prompt for a testing sample, $\bm{x}$, using a fine-to-coarse indexing strategy as, 
\begin{align} \label{eq:pre_p}
P_\theta(\mathcal{P}_{i}|\bm{x})=
\begin{cases}
1, & \mathrm{if} \ i=\left\lfloor\frac{\arg\max_i\left\{\theta_{ClaP}(\theta_{FeaE}(\bm{x}))_i\right\}}{|\mathcal{Y}_1|} \right\rfloor,\\
0,  & \mathrm{otherwise},
\end{cases} 
\end{align}
where $\lfloor\cdot\rfloor$ denotes the floor function, which rounds down to the nearest integer.\footnote{It holds when the tasks have an equal number of classes. For another CIL setting where there is a different number of classes in the initial task and an equal number of classes in the incremental tasks~\citep{masana2022class}, the index is simply adjusted to $\left\lfloor\frac{\arg\max_i\left\{\theta_{ClaP}(\theta_{FeaE}(\bm{x}))_i\right\}-|\mathcal{Y}_1|}{|\mathcal{Y}_2|} \right\rfloor+1$.}

\noindent
\textbf{Label prediction stage.} During this stage, we utilize the image, $\bm{x}$, again along with the selected prompt, $\bm{p}$, to predict the label, $y$.
\textit{In the training phrase}, the label classifier, $\theta_{ClaL}$, and the prompt, $\bm{p}$, are trained by,
\begin{align} \label{eq:clal}
\min\limits_{\left\{\theta_{ClaL}, \bm{p}\right\}} \left\{
    \mathcal{L}_l(\bm{x}, y)\equiv 
    -\sum_{j\in\mathcal{Y}_i}\delta(j,y)\log\left(\theta_{ClaL}\left(\theta_{FeaE}(\bm{x}, \bm{p})\right)_j\right)
\right\},
\end{align}
Note that the additional $\bm{p}$ is primarily precessed in MSA~\citep{wang2024hierarchical}, with two main implementations: prompt tuning (\textit{pro})~\citep{lester2021power} and prefix tuning (\textit{pre})~\citep{li2021prefix}.
Prompt tuning treats $\bm{p}$ as identical to $\bm{h}_Q$, $\bm{h}_K$ and $\bm{h}_V$, represented as:
\begin{align}
    \bm{h}^{pro} & = \mathrm{MSA}\left(Con(\bm{p}, \bm{h}_Q), Con(\bm{p}, \bm{h}_K), Con(\bm{p}, \bm{h}_V)\right). \notag
\end{align}
This approach increases the dimension of $\bm{h}^{pro}$ from $\mathbb{R}^{(HW/P^2+1)\times D}$ to $\mathbb{R}^{(HW/P^2+1+L)\times D}$ compared with $\bm{h}$ in \eqref{eq:h}.
In contrast, prefix tuning divides $\bm{p}$ equally into $\bm{p}_K\in\mathbb{R}^{L/2\times D}$ and $\bm{p}_V\in\mathbb{R}^{L/2\times D}$, applying them solely to $\bm{h}_K$ and $\bm{h}_V$ respectively:
\begin{align}
    \bm{h}^{pre} & = \mathrm{MSA}\left(\bm{h}_Q, Con(\bm{p}_K, \bm{h}_K), Con(\bm{p}_V, \bm{h}_V)\right),  \notag
\end{align}
where the output dimension remains consistent with that of $\bm{h}$.
\textit{In the inference phase}, we can predict the label for a test sample, $\bm{x}$, simply by, 
\begin{align} \label{eq:pre_y}
    P_{\theta}(y|\bm{p}, \bm{x})=\arg\max_i\left\{\theta_{ClaL}(\theta_{FeaE}(\bm{x}, \bm{p}))_i\right\}.
\end{align}
With these two predictions in \eqref{eq:pre_p} and \eqref{eq:pre_y}, we could achieve the prediction in \eqref{eq:p_yx}

\subsubsection{Feature translation} \label{subsub:fea}
Sequentially training classifiers on new tasks often introduces prediction bias towards the more recent classifier head, as historical data is unavailable to balance the older classifier heads within the global cross-entropy loss~\citep{huang2024etag}. 
To address this issue and achieve a balance between stability and plasticity in predictions, we adopt a feature translation strategy~\citep{petit2023fetril} to simultaneously train both the prompt-prediction and label-prediction classifiers.\footnote{Directly applying feature translation to a pre-trained model does not yield positive results, despite its success in conventional CIL frameworks built on ResNet; we validate this observation in \secref{subsec:abl}.}

As illustrated in \figref{fig:method}(b), during learning on the current task $t$, we first compute the mean feature (\ie prototype) for each new class by averaging the features within the class.
Next, we identify the nearest new prototype for each retained old prototype from previously learned tasks.
This step establishes a correspondence between old and new classes based on their prototype similarities, ensuring meaningful alignment.
Once the nearest new prototype is determined for each old prototype, we construct the old features by replacing the new prototype with the old one.
Specifically, we copy the new features and adjust them by subtracting their mean and adding the corresponding old prototype.
This translation process aligns the old features with the new feature space through parallel translation, ensuring that even if the prototypes are similar, a clear separation is maintained between their corresponding features.
The detailed feature translation algorithm is outlined in \algref{alg:fea_tra} in Appendix~\ref{app:fea}.

With the constructed old features, the losses in \eqref{eq:clap} and \eqref{eq:clal} are reformulated as follows:
\begin{align}
\begin{aligned}
    \mathcal{L}'_p(\bm{x}, y, \hat{\bm{f}}) \equiv -\sum_{j\in\mathcal{Y}_{1...t}} \delta(j,y) \log\left( \theta_{ClaP} \left( Con\left( \theta_{FeaE}( \bm{x}), \hat{\bm{f}} \right) \right)_j \right), 
\end{aligned}
\end{align}
\begin{align}
\begin{aligned}
    \mathcal{L}'_l(\bm{x}, y, \hat{\bm{f}}) \equiv  -\sum_{j\in\mathcal{Y}_{1...t}} \delta(j,y) \cdot \log\left( \theta_{ClaL} \left( Con\left(\theta_{FeaE}( \bm{x},\bm{p} ), \hat{\bm{f}} \right) \right)_j \right),
\end{aligned}
\end{align}
where $\hat{\bm{f}}$ is drawn from the set of constructed old features, $\{ \hat{\bm{f}}_{\mathcal{Y}_{i,j}} \mid i\in\{1,\cdots,t-1\}, j\in\{1,\cdots,|\mathcal{Y}_i|\}\}$; $\mathcal{Y}_{1...t}$ is the union of all classes up to task $t$, $\{ \bigcup_{i,j} \mathcal{Y}_{i,j} \mid i\in\{1,\cdots,t\}, j\in\{1,\cdots,\left|\mathcal{Y}_i\right|\} \}$. 
The trainable parameters and the inference phase in each stage are the same as those mentioned earlier in \secref{subsubsec:preprompt}.

\section{Experiment} \label{sec:exp}
\vspace{-5px}
\paragraph{Baseline methods.}
We compare our approach with three representative prompt-based methods: L2P~\citep{wang2022learning}, DualPrompt~\citep{wang2022dualprompt}, and CODA-Prompt~\citep{smith2023coda}, which serve as our main baselines.
L2P pioneers a dynamic prompt pool mechanism that retrieves task-specific prompts during inference, while DualPrompt introduces dual-level prompts (expert and shared) to decouple task-invariant and task-specific knowledge. 
CODA-Prompt further enhances prompt-based continual learning by dynamically composing prompts based on task correlations.
Additionally, we include Hide-Prompt \citep{wang2024hierarchical}, the most recent prompt-based continual learning method, which leverages hierarchical prompt distillation and adapter-based architecture to mitigate task interference and improve generalization.
We also incorporate S-Prompt++, an adaptation of the original domain-incremental learning method proposed by Wang \etal~\cite{wang2022s}, modified following \citep{wang2024hierarchical} to align with standard CIL methods by replacing task-specific multiple heads with a single shared classification head.
The output layer retains multiple heads associated.
Without specifying, we adopted the optimal training parameters and reported results from the original papers for all baseline methods to ensure a fair comparison.

\vspace{-5px}
\paragraph{Benchmark datasets.}
To ensure a meaningful evaluation, we avoid using ImageNet due to its overlap with the pretraining data of the
PTMs (e.g., ImageNet-21K~\citep{deng2009imagenet}). Instead, we follow~\citep{ijcai2024p924,wang2024hierarchical} and conduct experiments on four diverse benchmarks: CIFAR-100~\citep{krizhevsky2009learning}, ImageNet-R~\citep{hendrycks2021many}, CUB-200~\citep{wah2011caltech}, and 5-Datasets~\citep{hendrycks2021many}.
Specifically, we split \textbf{CIFAR-100} (100-class small-scale images) into 10 incremental tasks with disjoint classes, \textbf{ImageNet-R} (200-class large-scale images) into 10 incremental tasks with disjoint classes, and \textbf{CUB-200} (200-class fine-grained bird images) into 10 incremental tasks with disjoint classes. 
Additionally, \textbf{5-Datasets} comprises five distinct datasets, CIFAR-10~\citep{krizhevsky2009learning}, MNIST~\citep{lecun1998gradient}, Fashion-MNIST~\citep{xiao2017fashion}, SVHN~\citep{netzer2011reading}, and notMNIST~\citep{bulatov2011notmnist}, each treated as a separate incremental task to evaluate the robustness of methods under large inter-task distribution shifts.

\vspace{-5px}
\paragraph{Evaluation metrics.}
We evaluate our method using three widely adopted metrics: average accuracy~\citep{lopez2017gradient}, average incremental accuracy~\citep{rebuffi2017icarl}, and forgetting measure~\citep{chaudhry2018riemannian}.
Given the accuracy, $a_{k,j}\in[0, 1]$, evaluated on the $j$-th task ($j\le k$) after incrementally trained the model from tasks $1$ to $k$, the \textbf{average accuracy}, $A_T=\frac{1}{T}\sum_{j=1}^Ta_{T,j}$, reflects the final performance after learning $T$ tasks.
Building on this, the \textbf{average incremental accuracy},  $\bar{A}=\frac{1}{T}\sum_{k=1}^TA_k = \frac{1}{T}\sum_{k=1}^T\left(\frac{1}{k}\sum_{j=1}^ka_{k,j}\right)$, provides a absolute performance measure of continual learning.
In contrast, the \textbf{forgetting measure} defined as $F_T=\frac{1}{T-1}\sum_{j=1}^{T-1}f^T_j$, where $f^T_j=\max_{l\in\{{j,...,T-1}\}}a_{l,j}-a_{T,j}$, quantifies the relative performance measure for task $j$ by comparing the maximum accuracy ($\max_l a_{l,j}$) with its final accuracy ($a_{T,j}$). 
Higher $A_T\uparrow$ and $\bar{A}\uparrow$, coupled with lower $F_T\downarrow$, indicate superior performance.

\vspace{-5px}
\paragraph{Implementation details.}
We follow standard practices in prompt-based continual learning \citep{wang2022dualprompt, wang2024hierarchical} by employing a pretrained ViT-B/16 backbone~\citep{dosovitskiy2020image}. 
The model is optimized using Adam ($\beta_1=0.9$, $\beta_2=0.999$) with a batch size of $24$, trained for $50$ epochs per task to ensure convergence.
A key aspect of our configuration is the initial learning rate of $0.1$, which is intentionally set higher than typical values due to PrePrompt's separated prediction mechanism (empirically validated in ~\secref{subsec:abl}).
For prompt architecture, we implement a hierarchical design with $L=5$ trainable prompts inserted into transformer layers 1-5, consistent with established paradigms~\citep{wang2022dualprompt, wang2024hierarchical}.
All baselines follow their original implementations (detailed in Appendix~\ref{app:com}). Experiments were conducted on 8$\times$RTX 4090 GPUs (PyTorch 1.13~\citep{paszke2017automatic}) with three random seeds for statistical significance. Code available at \href{github.com/libo-huang/preprompt}{github.com/libo-huang/preprompt}.


\subsection{Overall compassion} \label{subsec:ove}
\vspace{-5px}
\paragraph{Performance comparison.}
We compare L2P~\citep{wang2022learning}, DualPrompt~\citep{wang2022dualprompt}, S-Prompt++~\cite{wang2022s}, CODA-Prompt~\citep{smith2023coda}, and HiDe-Prompt~\citep{wang2024hierarchical} and record the results in~\tabref{tab:com}.
The results demonstrate PrePrompt's consistent superiority across all benchmarks and metrics. On CIFAR-100, PrePrompt achieves $93.74\%$ $A_T$, outperforming the previous best (HiDe-Prompt's $92.61\%$) while reducing forgetting by $60\%$ ($1.27$ \vs $3.16$). This advantage extends to complex datasets: a 0.03-point improvement on ImageNet-R ($75.09\%$ \vs $75.06\%$) and significant gains on 5-Datasets ($94.54\%$ \vs $93.83\%$) and CUB-200 ($88.27\%$ \vs $86.56\%$). Notably, PrePrompt's $\bar{A}$ ($95.41\%$ on CIFAR-100) and $F_T$ metrics show it better balances stability and plasticity, maintaining high accuracy while minimizing catastrophic forgetting. The small standard deviations ($\pm0.15$ on CIFAR-100) further confirm its robustness. These improvements are particularly remarkable given the strong baselines: HiDe-Prompt's hierarchical decomposition and CODA-Prompt's dynamic combination. PrePrompt's success stems from its predictive prompting framework that inherently decouples task-specific and task-invariant learning, as validated by the ablation studies.

\newcommand{\RowsT}[1]{{\multirow{2}{*}{\begin{tabular}[c]{@{}c@{}}#1\end{tabular}}}}
\newcommand{\pms}[2]{%
    $#1\if\relax#2\relax\else{\scriptstyle\pm#2}\fi$
}
\CheckRmv{
\begin{table*}[tbp]
  \tabFormat
  \setlength{\tabcolsep}{2.1mm}
  \caption{Comprehensive performance comparison of prompt-based CIL methods across four benchmarks, showing final average accuracy ($A_T$), average incremental accuracy ($\bar{A}$), and forgetting measure ($F_T$). The top two results are distinguished by bold (best) and underline (second-best).  
  }\tabSpace
  \resizebox{0.998\linewidth}{!}{
  \begin{tabular}{l|ccccccccc} \toprule
    \RowsT{Method} 
    &  \multicolumn{3}{c}{CIFAR-100} & \multicolumn{3}{c}{ImageNet-R} & \multicolumn{2}{c}{5-Datasets} & CUB-200 \\ \cmidrule(lr){2-4} \cmidrule(lr){5-7} \cmidrule(lr){8-9} \cmidrule(lr){10-10}
    &  $A_T\uparrow$ & $\bar{A}\uparrow$ & $F_T\downarrow$   &   $A_T\uparrow$ & $\bar{A}\uparrow$ & $F_T\downarrow$   &   $A_T\uparrow$ & $F_T\downarrow$   & $A_T\uparrow$ \\ \cmidrule(lr){1-1} \cmidrule(lr){2-4} \cmidrule(lr){5-7} \cmidrule(lr){8-9} \cmidrule(lr){10-10}
    L2P~\citep{wang2022learning} & \pms{83.06}{0.17} & \pms{88.25}{0.01} & \pms{6.58}{0.40} & \pms{63.65}{0.12}	& \pms{67.25}{0.02} & \pms{7.51}{0.17}	&$81.84$ & $4.78$ & $74.48$ \\ 
    DualPrompt~\citep{wang2022dualprompt} & \pms{86.60}{0.19} & \pms{90.64}{0.01} & \pms{4.45}{0.16} & \pms{68.79}{0.31} & \pms{71.96}{0.04} & \pms{4.49}{0.14}	& $77.91$	& $13.11$ & $82.05$ \\ 
    S-Prompt++~\citep{wang2022s} & \pms{88.81}{0.18} & \pms{92.25}{0.03} & \pms{3.87}{0.05} & \pms{69.68}{0.12} & \pms{72.50}{0.04} & \pms{3.29}{0.05} & $86.06$ & $4.74$ & $82.08$ \\
    CODA-Prompt~\citep{smith2023coda} & \pms{86.94}{0.63} & \pms{91.57}{0.75} & \pms{4.04}{0.18} & \pms{70.03}{0.47} & \pms{74.26}{0.24} & \pms{5.17}{0.22} & $64.18$ & $17.23$ & $74.34$ \\
    HiDe-Prompt~\citep{wang2024hierarchical} & \pms{\underline{92.61}}{0.28} & \pms{\underline{94.03}}{0.01} & \pms{\underline{3.16}}{0.10} & \pms{\underline{75.06}}{0.12} & \pms{\underline{76.60}}{0.01} & \pms{\underline{2.17}}{0.19}	& \underline{$93.83$} & \underline{$0.44$} & \underline{$86.56$} \\
    PrePrompt & \pms{\bm{93.74}}{0.15} & \pms{\bm{95.41}}{0.11} & \pms{\bm{1.27}}{0.08} & \pms{\bm{75.09}}{0.08} & \pms{\bm{78.96}}{0.28} & \pms{\bm{1.11}}{0.00} & \bm{$94.54$} & \bm{$0.21$} & \bm{$88.27$} \\ \bottomrule
  \end{tabular}
    }
  \label{tab:com}  
\end{table*}
}

\vspace{-5px}
\paragraph{Complexity comparison.}
\begin{wraptable}{r}{0.55\textwidth}
\centering
\vspace{-6px}
\caption{Complexity comparison of prompt-based CIL methods on CIFAR-100. The top two results are distinguished by bold (best) and underline (second-best).}
\label{tab:volumn}
\tabSpace
\resizebox{0.998\linewidth}{!}{
    \begin{tabular}{l|lllcccc}\toprule
    Method & $\Delta{P}\downarrow$  & $\Delta{M}\downarrow$  & $A_T\uparrow$ \\ \midrule
    Finetune                                 & 0k        & 0 MB       & 81.21  \\
    L2P~\citep{wang2022learning}             & \textbf{138.24k}   & \textbf{0.527 MB} & 83.06    \\
    DualPrompt~\citep{wang2022dualprompt}    & 944.64k   & 3.604 MB  & 86.60   \\
    S-Prompt++~\citep{wang2022s}             & 1,543.68k & 6.035 MB   & 88.81  \\
    CODA-Prompt~\citep{smith2023coda}        & 3,840.00k & 14.648 MB  & 86.94  \\
    HiDe-Prompt~\citep{wang2024hierarchical} & 3,899.136k& 16.046 MB & \underline{92.61}   \\ 
    PrePrompt                                & \underline{384.00k}   & \underline{1.523 MB} & \textbf{93.74}    \\  
    \bottomrule
    \end{tabular}
    \vspace{-5px}
}
\end{wraptable}
The computational complexity of prompt-based CIL methods is primarily characterized by two key metrics: the additional training parameters ($\Delta{P}$) and memory overhead ($\Delta{M}$) required beyond the pretrained ViT backbone.
Since these complexity metrics scale with both the number of tasks and dataset size, we present our analysis based on the CIFAR-100 benchmark with 10 incremental tasks for consistent comparison.
As evidenced in \tabref{tab:volumn}, L2P demonstrates superior parameter efficiency with the lowest requirements ($138.24$k $\Delta{P}$ and $0.527$ MB $\Delta{M}$). 
This stands in marked contrast to more computationally intensive approaches like HiDe-Prompt and CODA-Prompt, which demand substantially greater resources ($3,899.136$k/$16.046$ MB and $3,840.00$k/$14.648$ MB, respectively).
Our proposed PrePrompt achieves an optimal balance between complexity and performance, requiring only $384.00$k parameters and $1.523$ MB memory while delivering state-of-the-art accuracy ($93.74\%$ $A_T$).
For the detailed computation across these two metrics, please refer to Appendix~\ref{app:bas}.

\subsection{Ablation study} \label{subsec:abl}

\vspace{-5px}
\paragraph{Effect of component.}
\begin{wraptable}{r}{0.5\textwidth}
    \centering
    \vspace{-18px}
    \caption{Ablation study of PrePrompt Framework on CIFAR-100 with 10 tasks. Key components are abbreviated as: P$_{pred.}$ for prompt prediction, P$_{f.t.}$ for the feature translation in prompt-prediction classifier, and L$_{f.t.}$ for the feature translation in label-prediction classifier.}
    \label{tab:ablation}
    \tabSpace
    \resizebox{0.998\linewidth}{!}{
    \begin{tabular}{l|ccc|cccc}   \toprule
    & P$_{pred.}$   & P$_{f.t.}$ & L$_{f.t.}$ & $A_T\uparrow$ & $\bar{A}\uparrow$ & $F_T\downarrow$  \\ \midrule  
    0  &             &            &            & 81.21 & 87.05 & 6.84 \\              
    1 &           &            & \checkmark & 82.25 & 87.47 & 5.08 \\   
    2 & \checkmark  &            &            & 90.31 & 93.43 & 2.37 \\   
    3 & \checkmark  & \checkmark &            & 90.55 & 93.55 & 2.13 \\            
    4 &  \checkmark  &            & \checkmark & 93.57 & 95.27 & 1.59 \\ 
    5 &  \checkmark  & \checkmark & \checkmark & 93.67 & 95.35 & 1.40 \\ \bottomrule   
    \end{tabular}
    }
    \vspace{-5px}
\end{wraptable}
\tabref{tab:ablation} demonstrates the effectiveness of our predictive prompting framework through three key findings.
First, direct feature translation on ViT shows limited gains (row 1 vs 0: $+1.04\%$ $A_T$).
This confirms that fixed pre-trained ViT naturally maintains cross-task feature comparability.
Second, our P$_{pred.}$ component alone achieves $90.31\%$ $A_T$ (row 2), proving the value of two-stage prediction. 
This establishes a task-aware feature space where translation becomes effective.
Third, in this enhanced space, L$_{f.t.}$ provides a significant $3.26\%$ $A_T$ boost (row 4 vs 2), while the full framework (row 5) reaches the peak performance of $93.67\%$ $A_T$.
The high $\bar{A}$ and low $F_T$ scores further confirm our method successfully balances stability and plasticity.
These results show that predictive prompting uniquely enables effective task feature alignment.

\begin{figure}[!t]
  \centering
  \begin{minipage}[t]{0.48\textwidth}
    \centering
    \includegraphics[width=\linewidth]{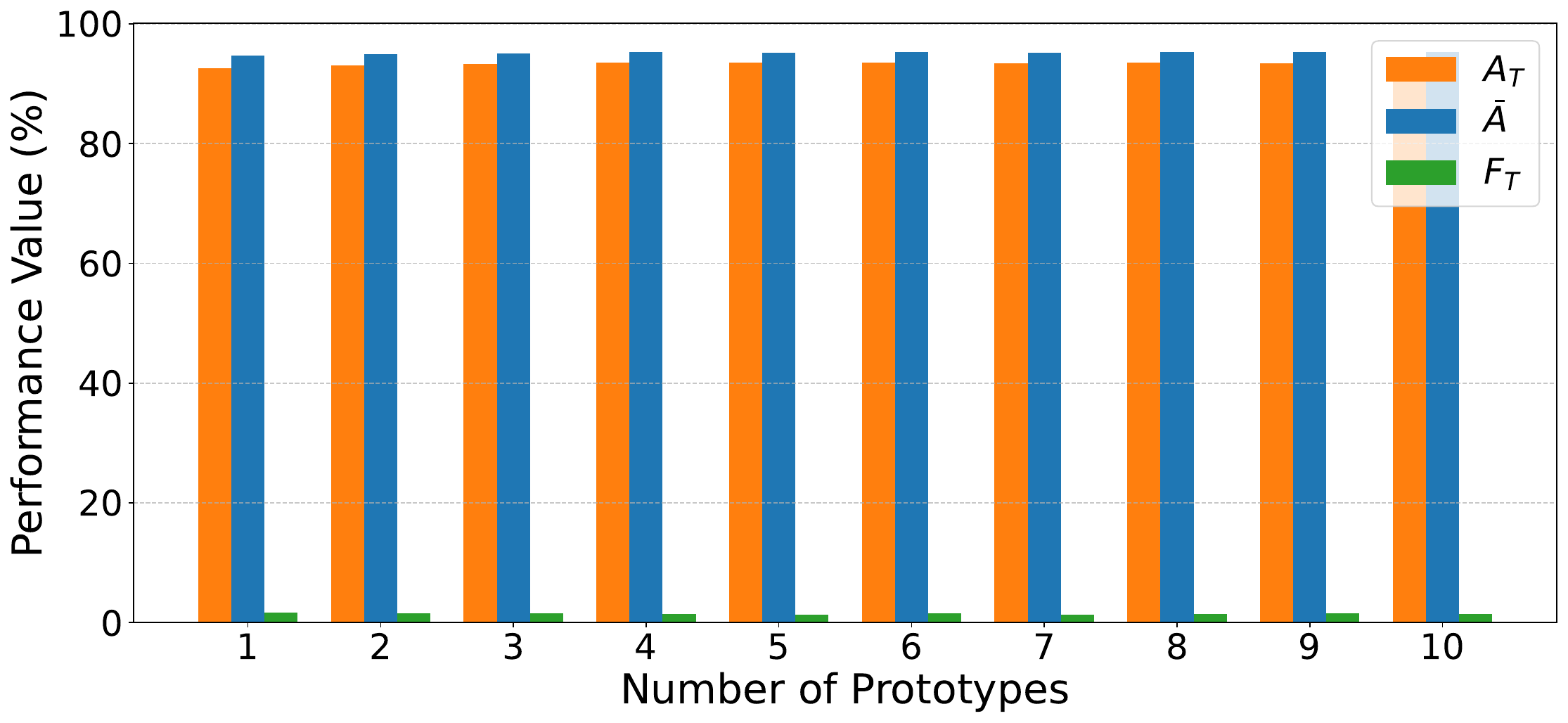}
    \vspace{-20px}
    \caption{Results of various prototype quantities.}
    \label{fig:centre}
  \end{minipage}
  \hfill
  \begin{minipage}[t]{0.48\textwidth}
    \centering
    \includegraphics[width=\linewidth]{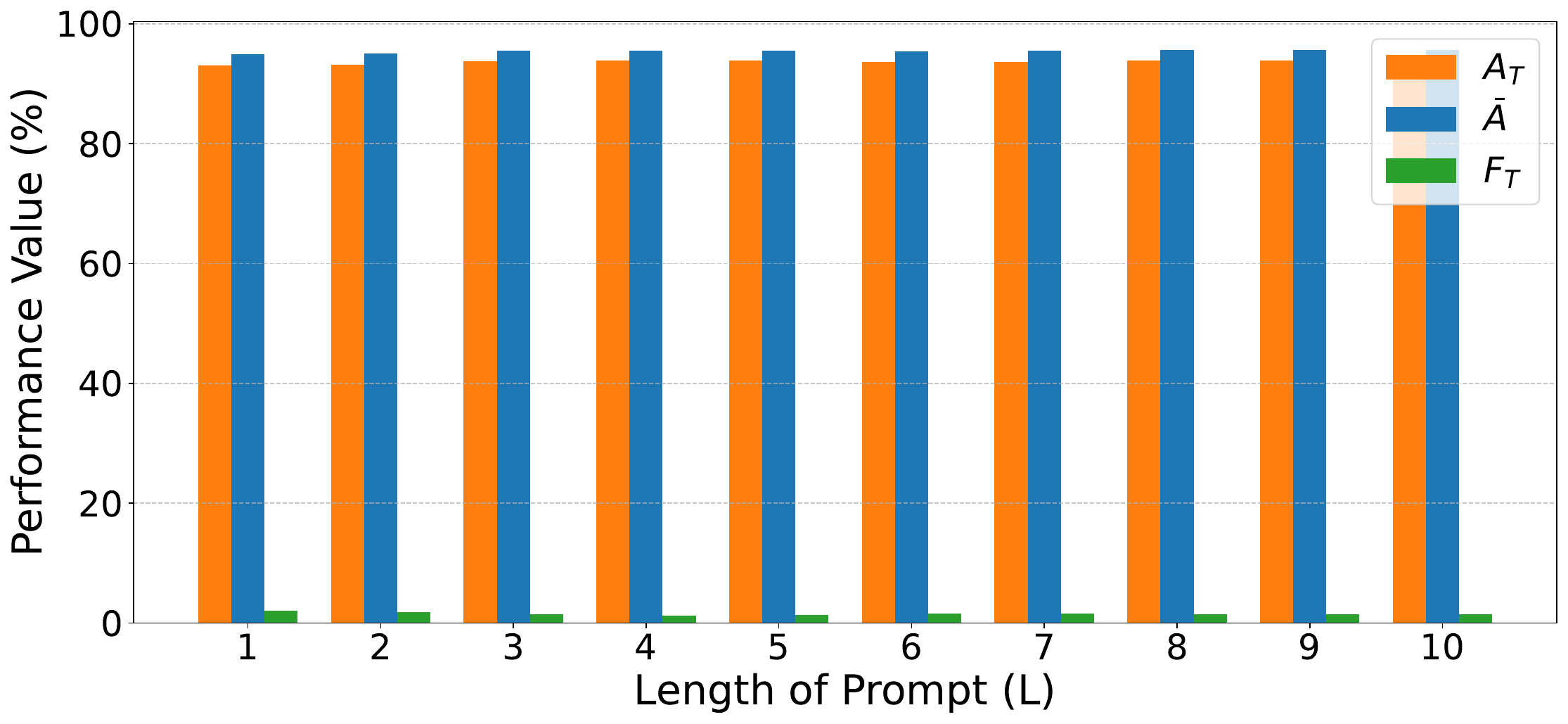}
    \vspace{-20px}
    \caption{Results of various prompt lengths.}
    \label{fig:length}
  \end{minipage}
  \vspace{-5px}
\end{figure}

\vspace{-5px}
\paragraph{Effect of the prototype quantities.}
As evidenced in \figref{fig:centre}, Preprompt exhibits remarkable robustness to the prototype quantities.
This stability could be attributed to two key factors: First, the predictive prompting mechanism inherently establishes well-separated cross-task feature representations, reducing the model's dependency on extensive prototype sets. Second, as detailed in \algref{alg:fea_tra}, the feature translation process effectively maintains parallel separation between old and new features through its prototype-agnostic alignment strategy. 

\vspace{-5px}
\paragraph{Effect of the prompt lengths.}
The results in \figref{fig:length} highlight Preprompt's robustness to the prompt lengths.
It maintains stable accuracy ($A_T$ between $93\%$ and $94\%$) for all lengths from 1 to 10.
This robustness stems from three key aspects of our design: (1) The pretrained ViT backbone's capability in accurately inferring task-specific prompt; (2) The integrated prompt efficiently propagates task information across transformer layers; and (3) The predictive prompting mechanism that dynamically adjusts feature extraction based on task requirements rather than relying on fixed prompt patterns. 
These innovations collectively make our framework remarkably resilient to prompt length variations, significantly reducing the engineering overhead typically associated with prompt tuning.

\begin{figure}[!t]
  \centering
  \small
  \setlength{\tabcolsep}{1pt}
  \begin{tabular}{cc}
    \includegraphics[width=0.48\textwidth]{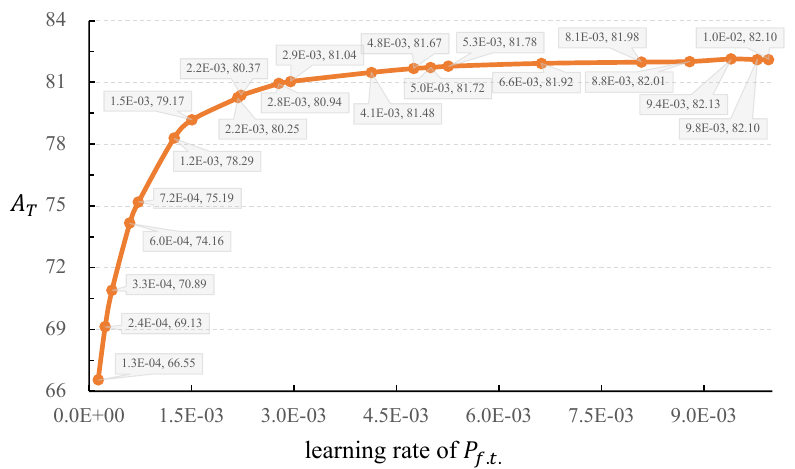} &
    \includegraphics[width=0.48\textwidth]{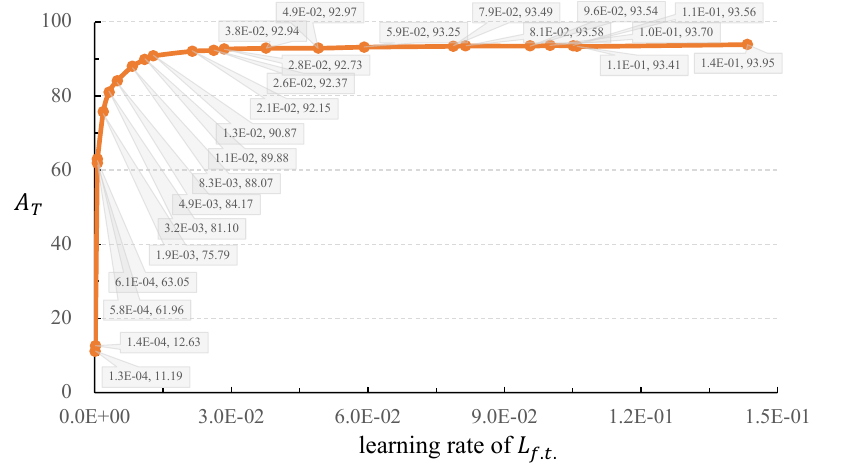} 
    \\
   (a) Sensitivity to the learning rate of P$_{f.t.}$  & (b) Sensitivity to the learning rate of L$_{f.t.}$ \\
  \end{tabular}
  \vspace{-5px}
  \caption{Results of various learning rates on CIFAR100 with 10 tasks.}
  \label{fig:lr}
  \vspace{-10px}
\end{figure}

\vspace{-5px}
\paragraph{Effect of learning rate.}
Benefiting from the unique prediction framework, PrePrompt is designed to be robust in selecting the learning rate.
We take the average accuracy ($A_T\uparrow$) as the performance indicator, as the prompt prediction is determined by a fixed indexing strategy as given in~\eqref{eq:pre_p}.
As evidenced in \figref{fig:lr}, both the prompt-prediction (P$_{f.t.}$) and label-prediction (L$_{f.t.}$) classifiers maintain strong performance across wide learning rate ranges ($1e-3$ to $1e-2$ and $0.05-0.14$, respectively), demonstrating the framework's stability.
Consistent conclusions are observed across multiple datasets (Appendix~\ref{app:mor}).
This tolerance stems from PrePrompt's three inherent properties: task-aware prompt prediction stabilizes gradients, feature translation provides implicit regularization through prototype alignment (\algref{alg:fea_tra}), and decoupled prediction stages prevent update conflicts. 
These advantages enable remarkably simple parameter tuning for PrePrompt.

\vspace{0px}
\section{Conclusion} \label{sec:con}
\vspace{-5px}
We present PrePrompt, a framework that advances prompt-based class incremental learning by leveraging pre-trained models' inherent classification ability to predict task-specific prompts. 
While existing methods struggle with limited prompt representation capacity, our approach combines decoupled prompt prediction with feature translation to dynamically balance stability and plasticity. 
Extensive experiments on four challenging benchmarks demonstrate that PrePrompt consistently outperforms state-of-the-art prompt-based CIL methods, achieving superior performance while requiring fewer additional parameters.
This work opens new directions for developing more efficient and scalable continual learning systems based on pre-trained models.

While PrePrompt demonstrates significant advantages, the current implementation has two main limitations: (1) reliance on task boundaries for feature translation, making it unsuitable for strict online learning, and (2) frozen backbone architecture that prevents knowledge refinement during continual adaptation. As fundamental research, the societal impact appears minimal, though ethical considerations should accompany future real-world applications. 


{
    \small
    \bibliography{main.bib}

\begin{thebibliography}{10}

\bibitem{bulatov2011notmnist}
Yaroslav Bulatov.
\newblock Notmnist dataset.
\newblock {\em Google (Books/OCR), Tech. Rep.[Online]. Available:
  http://yaroslavvb. blogspot. it/2011/09/notmnist-dataset. html}, 2:4, 2011.

\bibitem{cha2021co2l}
Hyuntak Cha, Jaeho Lee, and Jinwoo Shin.
\newblock Co2l: Contrastive continual learning.
\newblock In {\em Proceedings of the IEEE/CVF International Conference on
  Computer Vision}, pages 9516--9525, 2021.

\bibitem{chaudhry2018riemannian}
Arslan Chaudhry, Puneet~K Dokania, Thalaiyasingam Ajanthan, and Philip~HS Torr.
\newblock Riemannian walk for incremental learning: Understanding forgetting
  and intransigence.
\newblock In {\em Proceedings of the European conference on computer vision
  (ECCV)}, pages 532--547, 2018.

\bibitem{chaudhry2018efficient}
Arslan Chaudhry, Marc’Aurelio Ranzato, Marcus Rohrbach, and Mohamed
  Elhoseiny.
\newblock Efficient lifelong learning with a-gem.
\newblock In {\em International Conference on Learning Representations}, 2018.

\bibitem{deng2009imagenet}
Jia Deng, Wei Dong, Richard Socher, Li-Jia Li, Kai Li, and Li~Fei-Fei.
\newblock Imagenet: A large-scale hierarchical image database.
\newblock In {\em 2009 IEEE Conference on Computer Vision and Pattern
  Recognition}, pages 248--255. Ieee, 2009.

\bibitem{dosovitskiy2020image}
Alexey Dosovitskiy, Lucas Beyer, Alexander Kolesnikov, Dirk Weissenborn,
  Xiaohua Zhai, Thomas Unterthiner, Mostafa Dehghani, Matthias Minderer, Georg
  Heigold, Sylvain Gelly, et~al.
\newblock An image is worth 16x16 words: Transformers for image recognition at
  scale.
\newblock In {\em International Conference on Learning Representations}, 2020.

\bibitem{hendrycks2021many}
Dan Hendrycks, Steven Basart, Norman Mu, Saurav Kadavath, Frank Wang, Evan
  Dorundo, Rahul Desai, Tyler Zhu, Samyak Parajuli, Mike Guo, et~al.
\newblock The many faces of robustness: A critical analysis of
  out-of-distribution generalization.
\newblock In {\em Proceedings of the IEEE/CVF International Conference on
  Computer Vision}, pages 8340--8349, 2021.

\bibitem{huang2024etag}
Libo Huang, Yan Zeng, Chuanguang Yang, Zhulin An, Boyu Diao, and Yongjun Xu.
\newblock etag: Class-incremental learning via embedding distillation and
  task-oriented generation.
\newblock In {\em Proceedings of the AAAI Conference on Artificial
  Intelligence}, volume 38(11), pages 12591--12599, 2024.

\bibitem{jia2022visual}
Menglin Jia, Luming Tang, Bor-Chun Chen, Claire Cardie, Serge Belongie, Bharath
  Hariharan, and Ser-Nam Lim.
\newblock Visual prompt tuning.
\newblock In {\em European Conference on Computer Vision}, pages 709--727.
  Springer, 2022.

\bibitem{jung2023generating}
Dahuin Jung, Dongyoon Han, Jihwan Bang, and Hwanjun Song.
\newblock Generating instance-level prompts for rehearsal-free continual
  learning.
\newblock In {\em Proceedings of the IEEE/CVF International Conference on
  Computer Vision}, pages 11847--11857, 2023.

\bibitem{kirkpatrick2017overcoming}
James Kirkpatrick, Razvan Pascanu, Neil Rabinowitz, Joel Veness, Guillaume
  Desjardins, Andrei~A Rusu, Kieran Milan, John Quan, Tiago Ramalho, Agnieszka
  Grabska-Barwinska, et~al.
\newblock Overcoming catastrophic forgetting in neural networks.
\newblock {\em Proceedings of the National Academy of Sciences},
  114(13):3521--3526, 2017.

\bibitem{koller2009probabilistic}
Daphane Koller.
\newblock Probabilistic graphical models: Principles and techniques, 2009.

\bibitem{krizhevsky2009learning}
Alex Krizhevsky and Geoffrey Hinton.
\newblock Learning multiple layers of features from tiny images.
\newblock Technical Report~0, University of Toronto, Toronto, Ontario, 2009.

\bibitem{lecun2015deep}
Yann LeCun, Yoshua Bengio, and Geoffrey Hinton.
\newblock Deep learning.
\newblock {\em nature}, 521(7553):436--444, 2015.

\bibitem{lecun1998gradient}
Yann LeCun, L{\'e}on Bottou, Yoshua Bengio, and Patrick Haffner.
\newblock Gradient-based learning applied to document recognition.
\newblock {\em Proceedings of the IEEE}, 86(11):2278--2324, 1998.

\bibitem{lester2021power}
Brian Lester, Rami Al-Rfou, and Noah Constant.
\newblock The power of scale for parameter-efficient prompt tuning.
\newblock In {\em Proceedings of the 2021 Conference on Empirical Methods in
  Natural Language Processing}, pages 3045--3059, 2021.

\bibitem{li2021prefix}
Xiang~Lisa Li and Percy Liang.
\newblock Prefix-tuning: Optimizing continuous prompts for generation.
\newblock In {\em Proceedings of the 59th Annual Meeting of the Association for
  Computational Linguistics and the 11th International Joint Conference on
  Natural Language Processing (Volume 1: Long Papers)}, pages 4582--4597, 2021.

\bibitem{liu2020generative}
Xialei Liu, Chenshen Wu, Mikel Menta, Luis Herranz, Bogdan Raducanu, Andrew~D
  Bagdanov, Shangling Jui, and Joost~van de~Weijer.
\newblock Generative feature replay for class-incremental learning.
\newblock In {\em Proceedings of the IEEE/CVF Conference on Computer Vision and
  Pattern Recognition Workshops}, pages 226--227, 2020.

\bibitem{lopez2017gradient}
David Lopez-Paz and Marc'Aurelio Ranzato.
\newblock Gradient episodic memory for continual learning.
\newblock {\em Advances in Neural Information Processing Systems}, 30, 2017.

\bibitem{masana2022class}
Marc Masana, Xialei Liu, Bart{\l}omiej Twardowski, Mikel Menta, Andrew~D
  Bagdanov, and Joost Van De~Weijer.
\newblock Class-incremental learning: Survey and performance evaluation on
  image classification.
\newblock {\em IEEE Transactions on Pattern Analysis and Machine Intelligence},
  45(5):5513--5533, 2022.

\bibitem{mccloskey1989catastrophic}
Michael McCloskey and Neal~J Cohen.
\newblock Catastrophic interference in connectionist networks: The sequential
  learning problem.
\newblock In {\em Psychology of Learning and Motivation}, volume~24, pages
  109--165. Elsevier, 1989.

\bibitem{netzer2011reading}
Yuval Netzer, Tao Wang, Adam Coates, Alessandro Bissacco, Baolin Wu, Andrew~Y
  Ng, et~al.
\newblock Reading digits in natural images with unsupervised feature learning.
\newblock In {\em NIPS Workshop on Deep Learning and Unsupervised Feature
  Learning}, volume 2011(2), page~4. Granada, 2011.

\bibitem{paszke2017automatic}
Adam Paszke, Sam Gross, Soumith Chintala, Gregory Chanan, Edward Yang, Zachary
  DeVito, Zeming Lin, Alban Desmaison, Luca Antiga, and Adam Lerer.
\newblock Automatic differentiation in pytorch.
\newblock {\em Autodiff Workshop of Advances in Neural Information Processing
  Systems}, 2017.

\bibitem{petit2023fetril}
Gr{\'e}goire Petit, Adrian Popescu, Hugo Schindler, David Picard, and Bertrand
  Delezoide.
\newblock Fetril: Feature translation for exemplar-free class-incremental
  learning.
\newblock In {\em Proceedings of the IEEE/CVF Winter Conference on Applications
  of Computer Vision}, pages 3911--3920, 2023.

\bibitem{rebuffi2017icarl}
Sylvestre-Alvise Rebuffi, Alexander Kolesnikov, Georg Sperl, and Christoph~H
  Lampert.
\newblock icarl: Incremental classifier and representation learning.
\newblock In {\em Proceedings of the IEEE conference on Computer Vision and
  Pattern Recognition}, pages 2001--2010, 2017.

\bibitem{rusu2016progressive}
Andrei~A Rusu, Neil~C Rabinowitz, Guillaume Desjardins, Hubert Soyer, James
  Kirkpatrick, Koray Kavukcuoglu, Razvan Pascanu, and Raia Hadsell.
\newblock Progressive neural networks.
\newblock {\em arXiv preprint arXiv:1606.04671}, 2016.

\bibitem{serra2018overcoming}
Joan Serra, Didac Suris, Marius Miron, and Alexandros Karatzoglou.
\newblock Overcoming catastrophic forgetting with hard attention to the task.
\newblock In {\em International Conference on Machine Learning}, pages
  4548--4557. PMLR, 2018.

\bibitem{shin2017continual}
Hanul Shin, Jung~Kwon Lee, Jaehong Kim, and Jiwon Kim.
\newblock Continual learning with deep generative replay.
\newblock {\em Advances in Neural Information Processing Systems}, 30, 2017.

\bibitem{smith2023coda}
James~Seale Smith, Leonid Karlinsky, Vyshnavi Gutta, Paola Cascante-Bonilla,
  Donghyun Kim, Assaf Arbelle, Rameswar Panda, Rogerio Feris, and Zsolt Kira.
\newblock Coda-prompt: Continual decomposed attention-based prompting for
  rehearsal-free continual learning.
\newblock In {\em Proceedings of the IEEE/CVF Conference on Computer Vision and
  Pattern Recognition}, pages 11909--11919, 2023.

\bibitem{vaswani2017attention}
A~Vaswani.
\newblock Attention is all you need.
\newblock {\em Advances in Neural Information Processing Systems}, 2017.

\bibitem{wah2011caltech}
Catherine Wah, Steve Branson, Peter Welinder, Pietro Perona, and Serge
  Belongie.
\newblock The caltech-ucsd birds-200-2011 dataset.
\newblock Technical report, California Institute of Technology, 2011.

\bibitem{wang2024hierarchical}
Liyuan Wang, Jingyi Xie, Xingxing Zhang, Mingyi Huang, Hang Su, and Jun Zhu.
\newblock Hierarchical decomposition of prompt-based continual learning:
  Rethinking obscured sub-optimality.
\newblock {\em Advances in Neural Information Processing Systems}, 36, 2024.

\bibitem{wang2024comprehensive}
Liyuan Wang, Xingxing Zhang, Hang Su, and Jun Zhu.
\newblock A comprehensive survey of continual learning: Theory, method and
  application.
\newblock {\em IEEE Transactions on Pattern Analysis and Machine Intelligence},
  46(8):5362--5383, 2024.

\bibitem{wang2023attriclip}
Runqi Wang, Xiaoyue Duan, Guoliang Kang, Jianzhuang Liu, Shaohui Lin, Songcen
  Xu, Jinhu L{\"u}, and Baochang Zhang.
\newblock Attriclip: A non-incremental learner for incremental knowledge
  learning.
\newblock In {\em Proceedings of the IEEE/CVF Conference on Computer Vision and
  Pattern Recognition}, pages 3654--3663, 2023.

\bibitem{wang2022s}
Yabin Wang, Zhiwu Huang, and Xiaopeng Hong.
\newblock S-prompts learning with pre-trained transformers: An occam’s razor
  for domain incremental learning.
\newblock {\em Advances in Neural Information Processing Systems},
  35:5682--5695, 2022.

\bibitem{wang2022dualprompt}
Zifeng Wang, Zizhao Zhang, Sayna Ebrahimi, Ruoxi Sun, Han Zhang, Chen-Yu Lee,
  Xiaoqi Ren, Guolong Su, Vincent Perot, Jennifer Dy, et~al.
\newblock Dualprompt: Complementary prompting for rehearsal-free continual
  learning.
\newblock In {\em European Conference on Computer Vision}, pages 631--648.
  Springer, 2022.

\bibitem{wang2022learning}
Zifeng Wang, Zizhao Zhang, Chen-Yu Lee, Han Zhang, Ruoxi Sun, Xiaoqi Ren,
  Guolong Su, Vincent Perot, Jennifer Dy, and Tomas Pfister.
\newblock Learning to prompt for continual learning.
\newblock In {\em Proceedings of the IEEE/CVF Conference on Computer Vision and
  Pattern Recognition}, pages 139--149, 2022.

\bibitem{xiao2017fashion}
Han Xiao, Kashif Rasul, and Roland Vollgraf.
\newblock Fashion-mnist: a novel image dataset for benchmarking machine
  learning algorithms.
\newblock {\em arXiv preprint arXiv:1708.07747}, 2017.

\bibitem{xu2021artificial}
Yongjun Xu, Xin Liu, Xin Cao, Changping Huang, Enke Liu, Sen Qian, Xingchen
  Liu, Yanjun Wu, Fengliang Dong, Cheng-Wei Qiu, et~al.
\newblock Artificial intelligence: A powerful paradigm for scientific research.
\newblock {\em The Innovation}, 2(4), 2021.

\bibitem{zenke2017continual}
Friedemann Zenke, Ben Poole, and Surya Ganguli.
\newblock Continual learning through synaptic intelligence.
\newblock In {\em International Conference on Machine Learning}, pages
  3987--3995. PMLR, 2017.

\bibitem{zhou2024revisiting}
Da-Wei Zhou, Zi-Wen Cai, Han-Jia Ye, De-Chuan Zhan, and Ziwei Liu.
\newblock Revisiting class-incremental learning with pre-trained models:
  Generalizability and adaptivity are all you need.
\newblock {\em International Journal of Computer Vision}, pages 1--21, 2024.

\bibitem{ijcai2024p924}
Da-Wei Zhou, Hai-Long Sun, Jingyi Ning, Han-Jia Ye, and De-Chuan Zhan.
\newblock Continual learning with pre-trained models: A survey.
\newblock In Kate Larson, editor, {\em Proceedings of the Thirty-Third
  International Joint Conference on Artificial Intelligence, {IJCAI-24}}, pages
  8363--8371. International Joint Conferences on Artificial Intelligence
  Organization, 8 2024.
\newblock Survey Track.

\bibitem{zhou2024class}
Da-Wei Zhou, Qi-Wei Wang, Zhi-Hong Qi, Han-Jia Ye, De-Chuan Zhan, and Ziwei
  Liu.
\newblock Class-incremental learning: A survey.
\newblock {\em IEEE Transactions on Pattern Analysis and Machine Intelligence},
  2024.

\end{thebibliography}
}


\newpage
\appendix
\renewcommand{\thefigure}{A\arabic{figure}} 
\renewcommand{\thetable}{A\arabic{table}}   
\renewcommand{\thealgocf}{A\arabic{algocf}}
\setcounter{figure}{0}  
\setcounter{table}{0}   
\setcounter{algocf}{0}

{\centering\Large\bfseries
Appendix for \textit{
PrePrompt: Predictive prompting for class incremental learning
}\par}
\vspace{1em}

\section{More explanations}
This section provides more details about the variable definition, feature translation algorithm, and the baseline methods' details.
\subsection{Variable definition} \label{app:var}
\CheckRmv{
\begin{table*}[h]
    \tabFormat
    \setlength{\tabcolsep}{6pt}  
    \centering
    \caption{Variable definition.} \label{tab:var_def}
    \tabSpace
    \resizebox{0.998\linewidth}{!}{
    \begin{tabular}{lll}
    \toprule
    Variable   & Dimension & Definition \\
    \midrule  
    $H$, $W$      & -         & The resolution (height, width) of the original image \\            
    $C$          & -         & The number of image's channels ($3$ in common)  \\
    $P^2$      & -         & The resolution (height$\times$width) of each image patch \\
    $D$     & -     & The dimension of representation embedding \\
    $m$          & -         & The number of head in MSA  \\
    $L$ &   -   &  The sequence length of prompt    \\
    $\theta$   & -         & All of the parameters in pre-trained model \\
    \midrule
    $\bm{x}$        & $\mathbb{R}^{H\times W\times C}$              & Original image \\ 
    $y$    &   $\mathbb{R}$     &   Label of the image  \\
    $\bm{f}$     &   $\mathbb{R}^D$   &   Final feature representation    \\
    $\bm{x_s}$      & $\mathbb{R}^{HW/P^2\times P\times P\times C}$        & Sequence of image's patches  \\ 
    $\bm{cls}$                   & $\mathbb{R}^D$                           & Class token  \\
    $\theta_{posE}$                  & $\mathbb{R}^{(HW/P^2+1)\times D}$    & Position embedding \\
    $\bm{h}, \bm{h}', \bm{h}''$    &   $\mathbb{R}^{(HW/P^2+1)\times D}$   &   Sequence representations from the image patches and two adjacent attention blocks  \\
    $\bm{h}_{Q_i}, \bm{h}_{K_i}, \bm{h}_{V_i}$ & $\mathbb{R}^{(HW/P^2+1)\times D/m}$ & $i$-th (query, key, value) representation, where $i\in\{1,\cdots,m\}$ \\
    $\theta_{Q_i}, \theta_{K_i}, \theta_{V_i}$ & $\mathbb{R}^{D/m\times D/m}$ & $i$-th head (query, key, value) projection parameter matrices \\
    $\theta_{O}$            & $\mathbb{R}^{D\times D}$    & MultiHead projection parameter matrices \\
    $\bm{p}$ &   $\mathbb{R}^{L\times D}$    &   prompt \\
    \midrule
    $\theta_{ProjE}(\cdot)$        &       -       &       Embedding projector (transform patch from $\mathbb{R}^{P\times P\times C}$ to $\mathbb{R}^D$) \\
    $Con(\cdot,\cdot)$   & -             & Concatenation operation  \\
    $\theta_{FeaE}(\cdot)$  &   -   &   Feature extractor     \\
    $\theta_{ClaP}(\cdot)$      &   -   &  Classifier of prompt prediction    \\
    $\theta_{ClaL}(\cdot)$     &   -   &  Classifier of label prediction    \\
    \bottomrule
    \end{tabular}
    }
\end{table*}
}

\subsection{Feature translation algorithm} \label{app:fea}
\begin{algorithm*}[!th]
\caption{Feature translation on task $t$.}\label{alg:fea_tra}
\SetKwInOut{Input}{Input}
\SetKwInOut{Output}{Output}
\SetKwComment{Comment}{// }{}
\Input{Old prototypes of previously-learned $t-1$ tasks, $\mu_{\mathcal{Y}_{i,j}}$, $i\in\{1,...,t-1\}$, $j\in\{1,...,|\mathcal{Y}_i|\}$; 
new features of task $t$, $\bm{f}_{\mathcal{Y}_{t,k}}$, $k\in\{1,...,|\mathcal{Y}_{t}|\}$}
\Output{constructed old features of previously-learned $t-1$ tasks, $\hat{\bm{f}}_{\mathcal{Y}_{i,j}}$, $i\in\{1,...,t-1\}$, $j\in\{1,...,|\mathcal{Y}_i|\}$}
\For{$k\in\{1,...,|\mathcal{Y}_{t}|\}$}{
    $\mu_{\mathcal{Y}_{t,k}}=\mathrm{mean}(\bm{f}_{\mathcal{Y}_{t,k}})$
    \Comment*[r]{get the new prototype}
}
\For{$i\in\{1,...,t-1\}$}{
    \For{$j\in\{1,...,|\mathcal{Y}_i|\}$}{
        $k^*=\arg\min_k\{\mu_{\mathcal{Y}_{i,j}}-\mu_{\mathcal{Y}_{t,k}}\}, \ k\in\{1,...,|\mathcal{Y}_{t}|\}$ 
        \Comment*[r]{find the nearest new prototype}
        
        $\hat{\bm{f}}_{\mathcal{Y}_{i,j}}=\bm{f}_{\mathcal{Y}_{t,k^*}}-\mu_{\mathcal{Y}_{t,k^*}}+\mu_{\mathcal{Y}_{i,j}}$
        \Comment*[r]{construct the old features}
    }
}
\Return{$\hat{\bm{f}}_{\mathcal{Y}_{i,j}}$, $i\in\{1,...,t-1\}$, $j\in\{1,...,|\mathcal{Y}_i|\}$}
\end{algorithm*}

\subsection{Baselines methods' details} \label{app:bas}
All baseline methods are implemented following their original architectures as described in~\citep{wang2022dualprompt,wang2024hierarchical}, ensuring fair comparison under consistent experimental settings.
L2P~\citep{wang2022learning} employs a dynamic prompt pool consisting of $30$ key-value prompt pairs, with each key prompt having a length of $L=5$.
During training and inference, the top 5 most relevant prompts are selected.
DualPrompt~\citep{wang2022dualprompt} implements a dual-level prompt design: (1) general task-sharing prompts ($L=5$) in transformer layers 1-2 and (2) expert task-specific prompts ($L=20$) in transformer layers 3-5, using $10$ key-value pairs.
Building upon DualPrompt, S-Prompt++~\citep{wang2022s} extends the task-specific prompts to all transformer layers ($L=20$).
Both CODA-Prompt~\citep{smith2023coda} and HiDe-Prompt~\citep{wang2024hierarchical} follow similar layer-wise prompt insertion strategies (layers 1-5), with key differences in prompt composition: CODA-Prompt utilizes $100$ standalone prompts ($L=8$) without key-value pairing, while HiDe-Prompt introduces an auxiliary adapter layer for dynamic prompt selection.

\section{More experiments}
This section provides more experiments about the complexity analysis and the effect of the learning rate.
\subsection{Complexity details} \label{app:com}
The additional training parameters count ($\Delta{P}$) and memory overhead ($\Delta{M}$) beyond the pretrained ViT backbone are calculated as follows:
\begin{itemize}
\item
For L2P~\citep{wang2022learning}, the $30$ key-value prompt pairs (each key prompt having a length of $L=5$) with 768-dimensional embeddings require $30\times 5\times 768+30\times 768=138.240$k parameters ($0.527$ MB when stored as 32-bit floats).
\item
DualPrompt~\citep{wang2022dualprompt} combines $15,360$ parameters ($2\times 2\times 5\times 12\times 64$) from 2-layer general prompts ($L=5$) with $921,600$ expert value-prompt parameters ($3\times 2\times 10\times 20\times 12\times 64$) and 7,680 key-prompt parameters ($10\times 768$) from 2-layer general prompts ($L=20$), totaling $944.64$k parameters ($3.604$ MB). 
\item
S-Prompt++~\citep{wang2022s} extends this by deploying task-specific prompts across all 5 layers ($L=20$), yielding $1,543.68$ k prompt parameters ($5\times 2\times 10\times 20\times 12\times 64+10\times 768$) plus the stored $5$ centers for each task ($5\times 768\times 10=38,400$, $6.035$ MB total).
\item
CODA-Prompt~\citep{smith2023coda} takes $100$ composable prompts ($L=8$) with attentions in $3,840.00k$ parameters ($14.648$ MB).
\item
HiDe-Promptt~\citep{wang2024hierarchical} augments prompt parameters ($5\times 2\times 10\times 20\times 12\times 64=1,536,000$) with normalization layer and adapter layer parameters ($768+768+768\times 1536+1536+1536\times 768+768=2,363,136$), totaling $3,899.136$k parameters.
Besides, it stores $10$ mean and var centers for each learned task ($10\times 10\times 768\times 2\times 2$), consuming a total of $16.046$ MB memory overhead.
\item
PrePrompt achieves optimal efficiency with $384.00$k prompt parameters with $L=5$ ($5\times 2\times 10\times 5\times 12\times 64$) and stores $1$ prototype for each learned task ($1\times 10\times 768\times 2 = 15360$), consuming merely $1.523$ MB memory.
\end{itemize}
All calculations assume a 4-byte floating-point representation.

\begin{figure}[!t]
  \centering
  \small
  \setlength{\tabcolsep}{1pt}
  \begin{tabular}{cc}
    \includegraphics[width=0.51\textwidth]{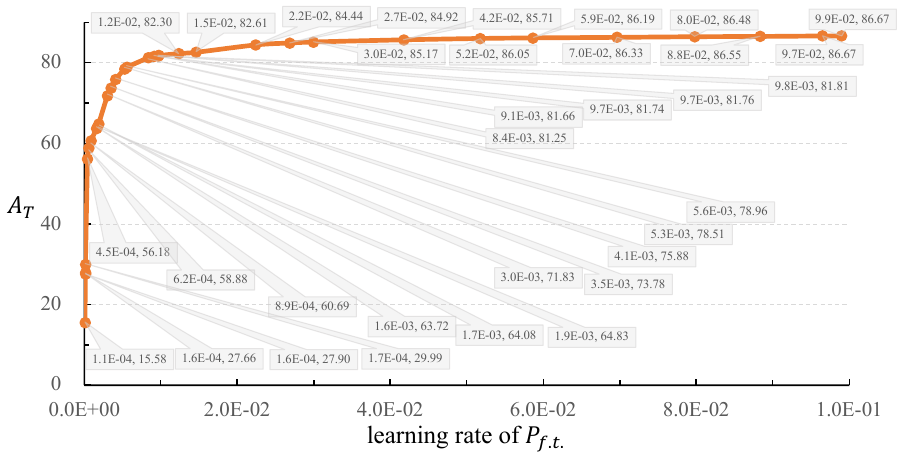} &
    \includegraphics[width=0.48\textwidth]{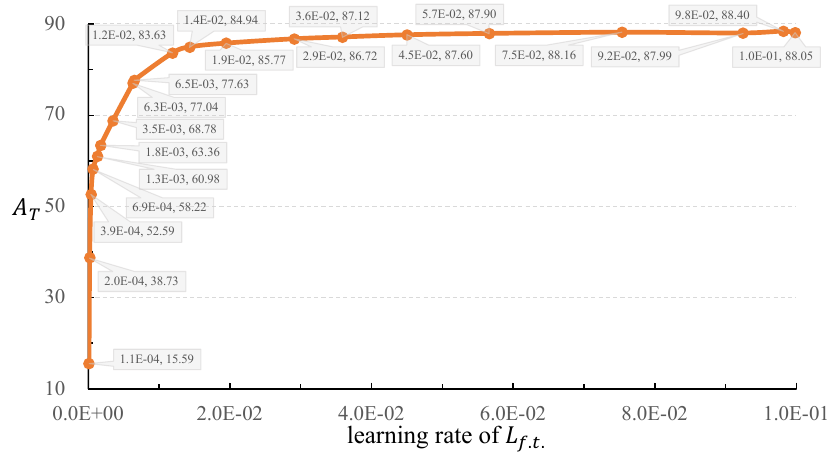} 
    \\
   (a) Sensitivity to the learning rate of P$_{f.t.}$  & (b) Sensitivity to the learning rate of L$_{f.t.}$ \\
  \end{tabular}
  \caption{Results of various learning rates on CUB-200 with 10 tasks.}
  \label{fig:lr_cub}
\end{figure}
\begin{figure}[!t]
  \centering
  \small
  \setlength{\tabcolsep}{1pt}
  \begin{tabular}{cc}
    \includegraphics[width=0.49\textwidth]{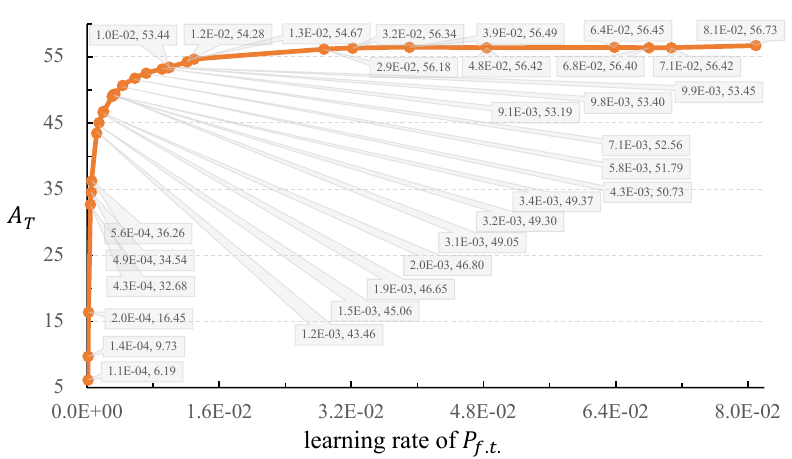} &
    \includegraphics[width=0.48\textwidth]{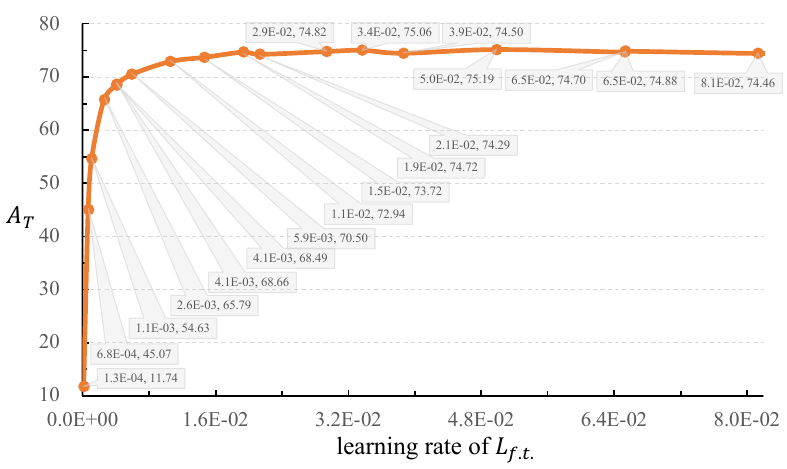} 
    \\
   (a) Sensitivity to the learning rate of P$_{f.t.}$  & (b) Sensitivity to the learning rate of L$_{f.t.}$ \\
  \end{tabular}
  \caption{Results of various learning rates on ImageNet-R with 10 tasks.}
  \label{fig:lr_imr}
\end{figure}
\begin{figure}[!t]
  \centering
  \small
  \setlength{\tabcolsep}{1pt}
  \begin{tabular}{cc}
    \includegraphics[width=0.48\textwidth]{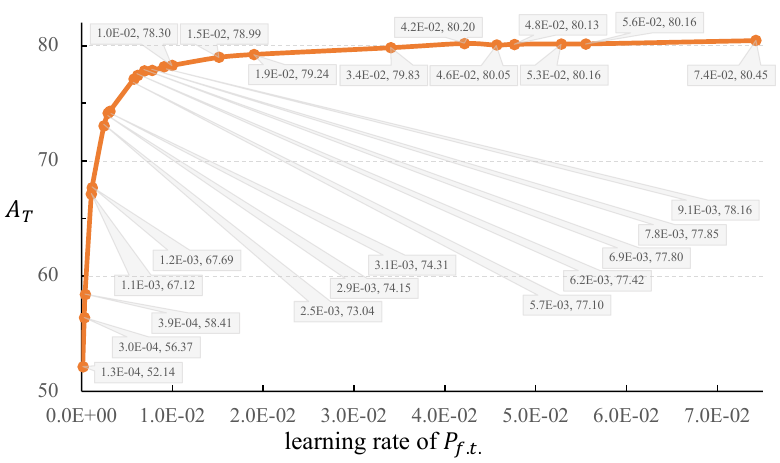} &
    \includegraphics[width=0.48\textwidth]{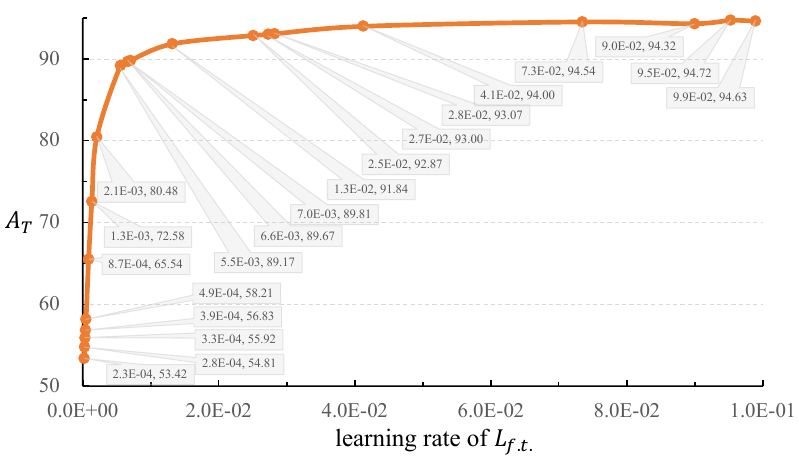} 
    \\
   (a) Sensitivity to the learning rate of P$_{f.t.}$  & (b) Sensitivity to the learning rate of L$_{f.t.}$ \\
  \end{tabular}
  \caption{Results of various learning rates on 5-Datasets with 5 tasks.}
  \label{fig:lr_5d}
\end{figure}
\subsection{More experience about the effect of learning rate} \label{app:mor}
Our framework's robustness to learning rate variations is further validated through comprehensive experiments on additional benchmarks. 
\figref{fig:lr_cub} demonstrates consistent performance patterns on CUB-200 (10-task), where both $P_{f.t.}$ and $L_{f.t.}$ maintain stable accuracy across their respective optimal ranges ($2e-2$ to $1e-1$ and $3e-2$ to $1e-1$). Similar trends are observed in~\figref{fig:lr_imr} and~\figref{fig:lr_5d} for ImageNet-R (10-task) and 5-Datasets (5-task), respectively, confirming the generalizability of our findings beyond CIFAR-100.

\section{Broader impacts} \label{app:bro}
PrePrompt advances responsible AI deployment by enabling efficient continual learning through predictive prompting. Our framework reduces computational costs while maintaining accuracy (\tabref{tab:ablation}), making CL more accessible for edge devices~\citep{zhou2024revisiting}. The feature translation mechanism (\secref{subsub:fea}) provides inherent privacy benefits by minimizing raw data retention during updates, addressing concerns in healthcare applications~\citep{petit2023fetril}.
We identify two key considerations: (1) The frozen backbone may perpetuate biases from pretraining data~\citep{ijcai2024p924}, particularly in sensitive domains like credit scoring; (2) Improved prompt efficiency could lower barriers for malicious adaptive systems~\citep{wang2024comprehensive}. We recommend complementary safeguards, including bias audits and detection frameworks for adversarial CL applications.

\section{License compliance statement} \label{app:lic}
This work complies with all relevant licenses for datasets and referenced implementations:

\begin{itemize}
\item CIFAR-100~\citep{krizhevsky2009learning}: Used under TorchVision's BSD-3 License. Source: \url{https://www.cs.toronto.edu/~kriz/cifar.html}

\item ImageNet-R~\citep{hendrycks2021many}: Licensed under MIT. Obtained from: \url{https://people.eecs.berkeley.edu/~hendrycks/imagenet-r.tar}

\item CUB-200~\citep{wah2011caltech}: Academic dataset from Caltech used under fair use principles. Source: \url{https://data.caltech.edu/records/65de6-vp158}

\item 5-Datasets (composite):
\begin{itemize}
\item CIFAR-10~\citep{krizhevsky2009learning}: BSD-3 (TorchVision)
\item MNIST~\citep{lecun1998gradient}: CC-BY-SA 3.0 (TorchVision)
\item Fashion-MNIST~\citep{xiao2017fashion}: MIT (TorchVision)
\item SVHN~\citep{netzer2011reading}: Non-commercial research only
\item notMNIST~\citep{bulatov2011notmnist}: Public Domain (CC0)
\end{itemize}

\item DualPrompt~\citep{wang2022dualprompt}: Apache 2.0 License. Code: \url{https://github.com/JH-LEE-KR/dualprompt-pytorch}

\item HiDe-Prompt~\citep{wang2024hierarchical}: MIT License. Code: \url{https://github.com/thu-ml/HiDe-Prompt}
\end{itemize}
All license terms have been respected, including: (a) Preservation of original copyright notices, (b) Compliance with usage restrictions (particularly for SVHN), and (c) Proper attribution in both documentation and source code.

\end{document}